\pdfoutput=1
\documentclass[11pt]{article}

\usepackage[final]{acl}

\usepackage{times}
\usepackage{latexsym}
\usepackage[T1]{fontenc}
\usepackage[utf8]{inputenc}
\usepackage{microtype}
\usepackage{inconsolata}
\usepackage{graphicx}
\usepackage{algorithm}
\usepackage{algorithmicx}
\usepackage{algpseudocode}
\usepackage{subcaption}
\usepackage{multirow}
\usepackage{rotating}
\usepackage{arydshln}
\usepackage{verbatim}
\usepackage{url}
\usepackage{amsmath}
\usepackage{amssymb}
\usepackage{mathtools}
\usepackage{hyperref}

\title{Quality Estimation Reranking for Document-Level Translation}

\author{
  Krzysztof Mrozinski \quad Minji Kang \quad Ahmed Khota \\
  {\bf Vincent Michael Sutanto} \quad {\bf Giovanni Gatti De Giacomo} \\
  Yaraku, Inc.
}

\begin{document}
\maketitle
\begin{abstract}
Quality estimation (QE) reranking is a form of quality-aware decoding which aims to improve machine translation (MT) by scoring and selecting the best candidate from a pool of generated translations. While known to be effective at the sentence level, its application to the increasingly prominent domain of document-level translation remains underexplored. In this work, we evaluate QE reranking performance on document-level (rather than the typical sentence-level) translation, using various learned and large language model (LLM)-based QE metrics. We find that with our best learned metric, SLIDE, BLEURT-20 scores improve by +2.00 with only two candidates, and by +5.09 with 32, across both decoder-only LLM models and encoder-decoder neural machine translation (NMT) models. Using the best LLM-based metric, GEMBA-DA, gains of +1.63 and +4.30 are achieved under the same conditions. Although gains shrink with longer inputs, reranking with 32 candidates yields improvements of +2.34 (SLIDE) and +1.40 (GEMBA-DA) on our longest documents (512-1024 source tokens). These findings demonstrate the practical value of document-level QE, with minimal runtime overhead given suitable translation models and hardware.
\end{abstract}

\section{Introduction}
\begin{figure}[!t]
    \centering
    \includegraphics[width=\linewidth]{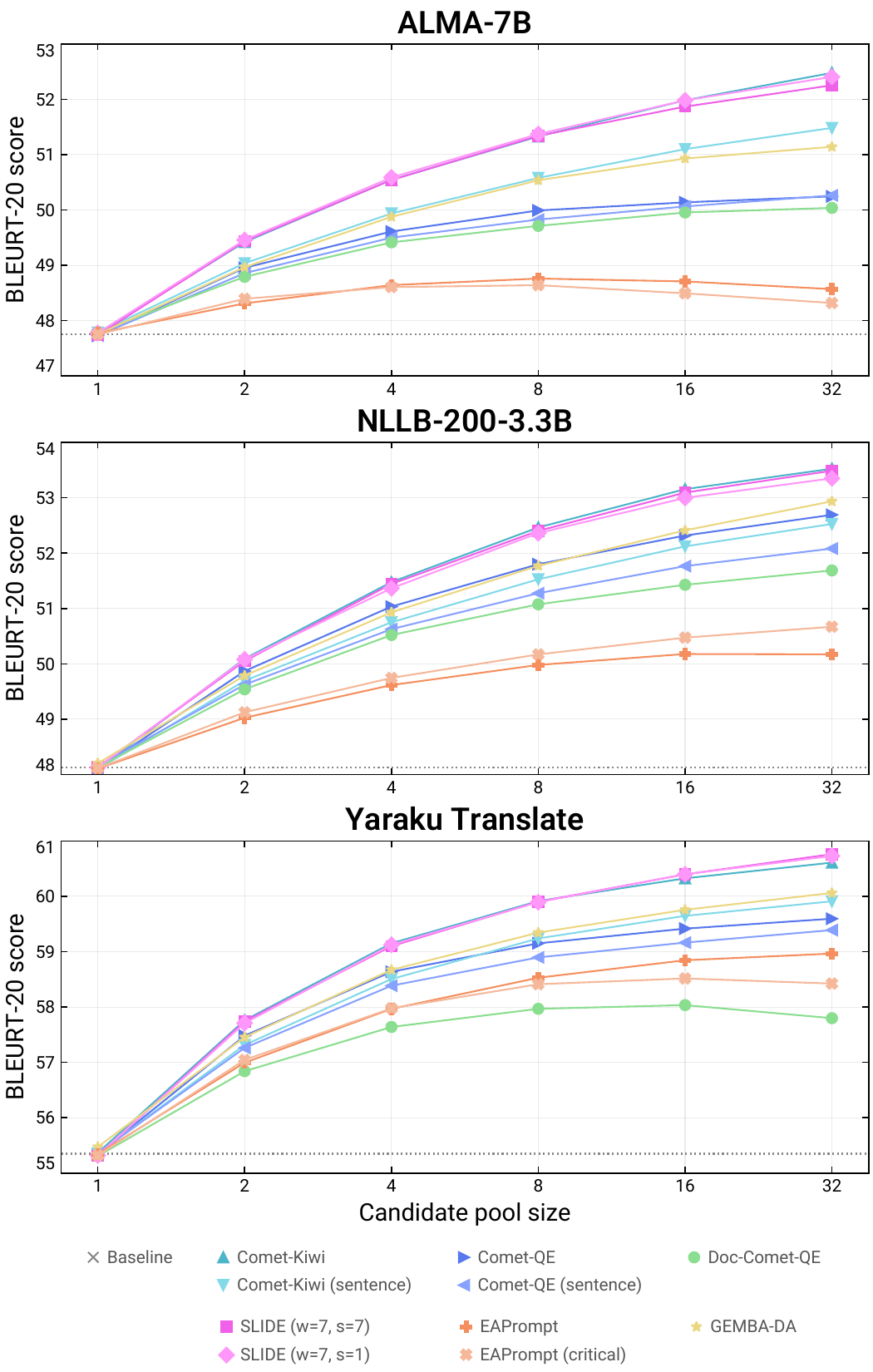}
    \caption{BLEURT-20 scores for QE reranking across different pool sizes, evaluated with all QE metrics and translation models. A pool size of 1 serves as the baseline (no reranking). Scores generally increase with larger pools under most QE metrics, for all translators.}
    \label{fig:pool}
\end{figure}

Machine Translation (MT) evaluation metrics are widely used to assess system performance, having been shown to align strongly with human evaluation \cite{ma-etal-2019-results}. In contrast, the standard decoding strategy of maximising model likelihood (MAP) has been shown to diverge from human evaluation \cite{eikema-aziz-2020-map, freitag2022highqualityhighmodel}. This motivates \textbf{quality-aware decoding} \cite{fernandes-etal-2022-quality}, where MT evaluation metrics are directly integrated into the translation process.

Quality-aware decoding uses an MT evaluation metric to select an optimal translation from a candidate pool. \textbf{Minimum Bayes-Risk} (MBR) decoding \cite{kumar-byrne-2004-minimum} uses reference-based metrics such as BLEU \cite{papineni-etal-2002-bleu} or COMET \cite{rei-etal-2020-comet}, comparing translation candidates against each other to select the highest utility candidate. Conversely, reference-free metrics, or Quality Estimation (QE) metrics, such as Comet-Kiwi \cite{rei-etal-2022-cometkiwi}, may be used for \textbf{QE reranking}, where the highest-scoring candidate is chosen as the final output. Although MBR mitigates weaknesses of MAP decoding \cite{muller-sennrich-2021-understanding}, it requires $O(N^2)$ pairwise comparisons compared to the linear $O(N)$ complexity of QE reranking, and translation quality gains over QE reranking are not definitive \cite{fernandes-etal-2022-quality, vernikos-popescu-belis-2024-dont}. Therefore, we focus our work on QE reranking.

While document-level MT is becoming increasingly prominent \cite{wang-etal-2024-benchmarking}, to date, document-level QE reranking has received little attention, despite one implementation showing promising results \cite{kudo-etal-2024-document}. The leading QE metrics such as Comet-Kiwi being sentence-level raises uncertainty about their suitability for document-level evaluation.

In this work, we investigate the applicability of QE metrics to document-level QE reranking. We evaluate translation quality improvements compared to standard decoding, examining differences across QE metrics, translation models, candidate pool sizes, and document lengths, as well as the associated computational trade-offs.

Our contributions are as follows:
\begin{itemize}
\item We demonstrate that QE reranking improves document-level MT quality across multiple QE metrics and translation models.
\item We analyse how reranking effectiveness varies with candidate pool size and document length.
\item We quantify the computational trade-offs of document-level QE reranking.
\end{itemize}

\section{Method}
\subsection{Translators}
For candidate generation, we evaluate both decoder-only large language models (LLMs) and encoder-decoder neural machine translation (NMT) models. While the use of NMT models is the traditional approach, the broad pretraining of LLMs typically enables them to generate more diverse outputs \cite{vernikos-popescu-belis-2024-dont}, making them well-suited for QE reranking.

We experiment with ALMA-7B \cite{xu2024paradigmshiftmachinetranslation}, a LLaMA2-7B LLM finetuned for translation, and NLLB-200-3.3B \cite{nllbteam2022languageleftbehindscaling}, a widely used multilingual NMT model. Although both were trained on sentence-level data, we found them sufficient for document-level translation, given the scarcity of publicly available document-trained models. For ALMA, the LLaMA pretraining and additional monolingual fine-tuning stage may further preserve document-level capabilities. We also evaluate Yaraku Translate, our proprietary NMT model trained directly for English-Japanese document translation.

For decoding, we adopt nucleus sampling ($p=0.9$) for ALMA \cite{touvron2023llama2openfoundation}. For NLLB, while beam search is the typical choice for NMT models, we opt for epsilon sampling ($\varepsilon=0.02$), shown to excel for MBR decoding \cite{freitag2023epsilonsamplingrocksinvestigating}. Temperatures of 0.6 (ALMA) and 0.5 (NLLB) balance candidate pool diversity and quality, and have been effective for QE reranking \cite{vernikos-popescu-belis-2024-dont}. For Yaraku Translate, we opt for diverse beam search \cite{vijayakumar2018diversebeamsearchdecoding} as an alternative to sampling, to address difficulties in achieving diverse yet high-quality outputs with sampling. We set $G=16$ groups and $\lambda=0.5$ diversity strength.

\subsection{QE Metrics}

\subsubsection{Learned QE Metrics}

We adopt the COMET model family as baseline QE metrics, namely \textbf{COMET-QE} \cite{rei-etal-2020-unbabels} and \textbf{Comet-Kiwi} \cite{rei-etal-2022-cometkiwi}. Although trained for sentence-level evaluation, we test two strategies for adapting them to the document-level.

The first averages sentence-level predictions across a document. Documents are segmented into sentences using Punkt \cite{kiss-strunk-2006-unsupervised} for English and a simple regular expression for Japanese, then aligned by order. When source and target sentence counts differ, the shorter text is padded by duplicating its final sentence, ensuring equal segment counts so that all sentences are scored. While this approach closely aligns with the intended use case, it is flawed in practice since document-level translators rarely preserve one-to-one sentence alignment, which compounds with longer documents. We refer to these metrics as \textbf{COMET-QE (sentence)} and \textbf{Comet-Kiwi (sentence)}.

The second strategy passes full documents as single segments. Although not the intended use-case, prior work shows this may perform comparably to metrics directly trained for longer-context evaluation \cite{deutsch-etal-2023-training}, likely due to the long-context pretraining of the underlying InfoXLM encoders \cite{chi-etal-2021-infoxlm}. This method is constrained by sequence length limits (512 tokens for source + target for Comet-Kiwi, 512 per text for COMET-QE). We refer to these metrics simply as \textbf{COMET-QE} and \textbf{Comet-Kiwi}.

\textbf{Doc-COMET-QE} \cite{vernikos-etal-2022-embarrassingly} extends COMET-QE by concatenating two preceding source and target sentences (where available) to provide additional document-level context. Sentence-level score is calculated only for the current sentence via masking, and document-level score as the average of all sentence-level scores. This is compatible with COMET-QE, which pools token representations and allows for selective masking, but not with Comet-Kiwi, whose representation collapses into a single \texttt{[CLS]} token. Although shown to improve accuracy over COMET-QE in isolated evaluations, Doc-COMET-QE inherits the same alignment problems as COMET-QE (sentence), limiting its utility for QE reranking. We use the same sentence alignment and padding approaches as outlined in COMET-QE (sentence).

\textbf{SLIDE} \cite{raunak-etal-2024-slide} is a document-level QE approach requiring no architectural changes, so we implement it on top of Comet-Kiwi. It segments documents into fixed-sentence-width, strided windows, scoring each window independently and averaging to obtain a document-level score. SLIDE is identical to Comet-Kiwi for any documents shorter than the window length but mitigates sequence length limitations for longer documents. The original work reported optimal performance with $w=6$, $s=6$ (window size, stride) in idealised conditions where documents segmented evenly. We instead adopt their proposed weighted partial window approach, which accommodates arbitrary document lengths. Since the best configuration is unclear, we experiment with both $w=7$, $s=7$ and $w=7$, $s=1$ as they both show good performance while representing two extremes of the method. We use the same padding approach as outlined in COMET-QE (sentence).

\subsubsection{LLM-based QE Metrics}

Given their strong performance in translation tasks, LLMs are a natural choice for document-level QE. We evaluate two prompting-based methods using Gemma 3 27B \cite{gemmateam2025gemma3technicalreport} as the backbone. Although originally developed for sentence-level QE, we expect them to transfer effectively to documents due to the long-context capabilities of LLMs in MT \cite{karpinska-iyyer-2023-large}.

\textbf{GEMBA-DA} \cite{kocmi-federmann-2023-large} tasks the LLM with Direct Assessment, assigning a score from 0 to 100 in a zero-shot manner. This method is efficient, requiring minimal token generation. We make minimal modifications to the original prompt to account for Gemma being instruction tuned. To mitigate the unpredictable nature of LLM output, we adopt the failure-recovery strategy from the original work, retrying with gradually higher temperature for up to five attempts, after which the candidate is discarded. This introduces a small chance of no valid candidates kept, so a fallback QE metric may be important.

\textbf{EAPrompt} \cite{lu-etal-2024-error} emulates the MQM human evaluation framework \cite{freitag-etal-2021-experts}. To compute the score, the LLM identifies major and minor errors, from which a weighted sum is computed via a regular expression. We weight major errors eight times higher than minor errors, shown to be effective for segment-level evaluation. We adopt one-shot prompting with language-pair-specific in-context examples, with minor prompt adjustments to accommodate the output style of Gemma. Both EAPrompt and GEMBA-DA frequently produce tied scores, which we resolve via random selection.

Although EAPrompt has shown state-of-the-art (SOTA) performance, our experiments revealed some limitations. First, unlike GEMBA-DA, it lacks a failure-recovery mechanism: erroneous outputs with no listed errors are indistinguishable from valid assessments of perfect translations. Second, the scoring scheme is overly lenient on critical translation errors—e.g., a nonsensical translation may only receive one major error, while a flawed yet comprehensible translation has more scope for identifying several errors. This issue is likely amplified at the document level, where long contexts increase the risk of critical errors. To address this, we introduce \textbf{EAPrompt-Critical}, which adds a critical error category weighted at 100. 

\section{Experiments}
For our main experiment, we generate a large pool of candidate translations for each source text in our dataset, score them with each QE metric, and then trim the candidate pool to various sizes. From each pool, we select the top-scoring candidate and evaluate it against the reference translation using several reference-based metrics.

\subsection{Dataset}

\begin{figure}[t]
    \centering
    \includegraphics[width=\linewidth]{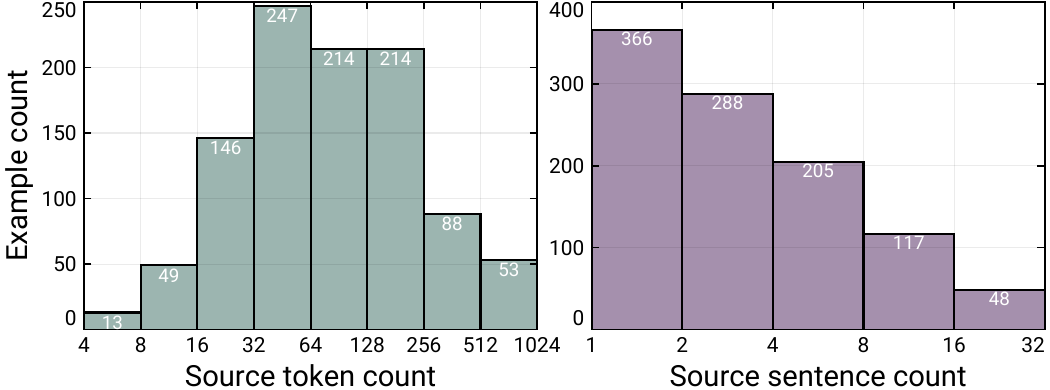}
    \caption{Distribution of source token and source sentence counts across our WMT23 dataset. Average example source text is 4.30 sentences and 138 tokens long.}
    \label{fig:dataset}
\end{figure}

We use the WMT23 test set \cite{freitag-etal-2023-results} as our source of ground-truth translations, evaluating bidirectionally between English and Japanese. As the dataset is segmented at the document, paragraph, and sentence levels, we merge segments to obtain document-level translations. We augment the data with a balanced mix of full documents and individual paragraphs to better examine the relationship between document length and performance. As the dataset was released after the COMET models we evaluate, there is no risk of overfitting, although some risk remains for Gemma. Dataset length distributions are shown in \autoref{fig:dataset}.

\subsection{Evaluation Metrics}

We evaluate QE reranking performance using reference-based metrics, treating the reranked translation as the hypothesis and the dataset translation as the reference. We consider two families: \textbf{neural metrics}, using \textbf{BLEURT-20} \cite{sellam-etal-2020-bleurt} and \textbf{COMET-22} \cite{rei-etal-2022-comet}, the latter setting the SOTA in WMT 2022 shared task \cite{freitag-etal-2022-results}; and \textbf{LLM-based evaluation}, using the reference-based prompting framework \textbf{GEMBA-DA} \cite{kocmi-federmann-2023-large}, which we had found to outperform EAPrompt. For the backbone LLM, we use GPT-4.1-mini \cite{openai2024gpt4technicalreport} for its strong natural language understanding capabilities and to minimise the risk of overfitting with our QE metrics using Gemma. Notably, this is the only evaluation metric directly compatible with document-level translation, as both COMET and BLEURT impose a strict 512 token cap, limiting their reliability on long documents. 

We acknowledge the risk of overfitting when using the same metric family for both QE and evaluation, which can lead to evaluation scores diverging from human judgement \cite{fernandes-etal-2022-quality}. Nonetheless, we include COMET-22 as an evaluation metric to enable comparison across a broad range of evaluators and discuss the implications of overfitting in \autoref{sec:Section 4}.

\section{Results}
\label{sec:Section 4}
{
\setlength{\tabcolsep}{2.9pt}
\begin{table*}[t]
    \centering
    \scriptsize
    \newcommand{\Hi}[2]{\begingroup\setlength{\fboxsep}{0pt}\colorbox{#1}{\strut #2}\endgroup}
    \definecolor{bgc}{HTML}{FEF0C9}
    \definecolor{bbc}{HTML}{D5E6CF}
    \definecolor{bcc}{HTML}{D3CCE4}
    \newcommand{\bg}[1]{\textbf{\Hi{bgc}{#1}}}
    \newcommand{\bb}[1]{\textbf{\Hi{bbc}{#1}}}
    \newcommand{\bc}[1]{\textbf{\Hi{bcc}{#1}}}
    \newcommand{\basea}{{\bg{47.39} / \bb{47.77} / \bc{72.62}}}
    \newcommand{\basen}{{\bg{50.92} / \bb{48.09} / \bc{73.72}}}
    \newcommand{\basey}{{\bg{70.21} / \bb{55.37} / \bc{78.61}}}

    \begin{tabular}{l|c|c|c|c|c|c}
    \hline
    \rule{0pt}{3ex} \textbf{QE Metric} & \textbf{Baseline} & \textbf{2} & \textbf{4} & \textbf{8} & \textbf{16} & \textbf{32} \\
    \hline
    \multicolumn{7}{c}{\rule{0pt}{3ex} \textbf{ALMA-7B} (\Hi{bgc}{GEMBA-DA} / \Hi{bbc}{BLEURT-20} / \Hi{bcc}{COMET-22})} \\
    \hline
    Comet-Kiwi             & \multirow{10}{*}{\basea} & 51.24 / 49.42 / \bc{74.81}      & 53.77 / 50.54 / \bc{76.04}      & 55.46 / 51.33 / 76.80           & 56.91 / 51.98 / 77.35           & 58.15 / \bb{52.48} / 77.71     \\
    Comet-QE               &                          & 49.88 / 48.95 / 74.79           & 51.32 / 49.61 / 75.91           & 52.06 / 49.99 / 76.52           & 52.21 / 50.14 / 76.83           & 52.47 / 50.25 / 77.05          \\
    Comet-Kiwi (sentence)  &                          & 50.18 / 49.04 / 74.16           & 52.12 / 49.94 / 75.04           & 53.57 / 50.58 / 75.64           & 54.68 / 51.10 / 76.07           & 55.57 / 51.48 / 76.37          \\
    Comet-QE (sentence)    &                          & 49.39 / 48.86 / 74.45           & 50.62 / 49.50 / 75.32           & 51.18 / 49.83 / 75.73           & 51.31 / 50.06 / 75.85           & 51.22 / 50.26 / 75.85          \\
    Doc-Comet-QE           &                          & 49.26 / 48.79 / 74.34           & 50.50 / 49.41 / 75.28           & 50.92 / 49.71 / 75.74           & 51.16 / 49.95 / 75.98           & 51.10 / 50.04 / 76.05          \\
    SLIDE (w=7, s=7)       &                          & 51.24 / 49.44 / 74.74           & 53.66 / 50.54 / 75.95           & 55.50 / 51.34 / 76.75           & 56.77 / 51.87 / 77.29           & 58.17 / 52.25 / 77.74          \\
    SLIDE (w=7, s=1)       &                          & 51.23 / \bb{49.46} / 74.76      & 53.84 / \bb{50.59} / 76.03      & 55.61 / \bb{51.37} / \bc{76.82} & 56.99 / \bb{51.98} / \bc{77.41} & 58.32 / 52.40 / \bc{77.81}     \\
    GEMBA-DA               &                          & \bg{51.95} / 48.97 / 74.14      & \bg{55.44} / 49.88 / 75.25      & \bg{57.96} / 50.54 / 75.87      & \bg{59.64} / 50.93 / 76.30      & \bg{60.81} / 51.14 / 76.54     \\
    EAPrompt               &                          & 49.18 / 48.31 / 72.88           & 50.40 / 48.64 / 72.76           & 51.15 / 48.76 / 72.40           & 51.42 / 48.71 / 71.88           & 51.53 / 48.57 / 71.30          \\
    EAPrompt (critical)    &                          & 49.53 / 48.40 / 73.29           & 50.80 / 48.61 / 73.29           & 51.39 / 48.64 / 73.00           & 51.58 / 48.49 / 72.46           & 51.54 / 48.32 / 71.88          \\
    \hline
    \multicolumn{7}{c}{\rule{0pt}{3ex} \textbf{NLLB-200-3.3B} (\Hi{bgc}{GEMBA-DA} / \Hi{bbc}{BLEURT-20} / \Hi{bcc}{COMET-22})} \\
    \hline
    Comet-Kiwi             & \multirow{10}{*}{\basen} & 55.88 / \bb{50.09} / \bc{75.94} & 58.94 / \bb{51.48} / \bc{77.26} & 61.12 / \bb{52.46} / \bc{78.14} & 62.66 / \bb{53.16} / \bc{78.78} & 63.75 / \bb{53.53} / \bc{79.25}\\
    Comet-QE               &                          & 54.47 / 49.87 / 75.93           & 56.53 / 51.03 / 77.19           & 57.67 / 51.80 / 78.03           & 58.31 / 52.32 / 78.63           & 58.65 / 52.69 / 79.06          \\
    Comet-Kiwi (sentence)  &                          & 54.64 / 49.70 / 75.35           & 57.08 / 50.75 / 76.32           & 58.86 / 51.53 / 76.97           & 60.22 / 52.12 / 77.49           & 61.36 / 52.53 / 77.91          \\
    Comet-QE (sentence)    &                          & 53.98 / 49.62 / 75.52           & 55.97 / 50.63 / 76.55           & 57.12 / 51.28 / 77.24           & 58.00 / 51.77 / 77.76           & 58.58 / 52.08 / 78.07          \\
    Doc-Comet-QE           &                          & 53.70 / 49.54 / 75.40           & 55.25 / 50.52 / 76.27           & 55.99 / 51.08 / 76.79           & 56.26 / 51.43 / 77.07           & 56.13 / 51.69 / 77.22          \\
    SLIDE (w=7, s=7)       &                          & 55.63 / 50.05 / 75.84           & 58.68 / 51.44 / 77.10           & 60.79 / 52.40 / 77.90           & 62.26 / 53.09 / 78.51           & 63.27 / 53.49 / 78.91          \\
    SLIDE (w=7, s=1)       &                          & 55.73 / 50.08 / 75.90           & 58.70 / 51.37 / 77.15           & 60.82 / 52.36 / 78.01           & 62.26 / 53.00 / 78.65           & 63.17 / 53.35 / 79.06          \\
    GEMBA-DA               &                          & \bg{56.07} / 49.79 / 75.49      & \bg{59.57} / 50.93 / 76.49      & \bg{62.20} / 51.77 / 77.21      & \bg{64.04} / 52.41 / 77.71      & \bg{65.55} / 52.94 / 78.10     \\
    EAPrompt               &                          & 53.18 / 49.03 / 74.20           & 54.56 / 49.62 / 74.33           & 55.25 / 49.98 / 74.08           & 55.57 / 50.18 / 73.65           & 55.42 / 50.17 / 72.97          \\
    EAPrompt (critical)    &                          & 53.51 / 49.12 / 74.56           & 55.13 / 49.75 / 74.90           & 56.04 / 50.17 / 74.99           & 56.65 / 50.47 / 74.95           & 56.92 / 50.67 / 74.80          \\
    \hline
    \multicolumn{7}{c}{\rule{0pt}{3ex} \textbf{Yaraku Translate} (\Hi{bgc}{GEMBA-DA} / \Hi{bbc}{BLEURT-20} / \Hi{bcc}{COMET-22})} \\
    \hline
    Comet-Kiwi             & \multirow{10}{*}{\basey} & 76.21 / \bb{57.76} / \bc{80.66} & 79.33 / \bb{59.15} / 81.80     & 80.81 / \bb{59.91} / \bc{82.45} & 81.49 / 60.32 / 82.84            & 81.81 / 60.61 / 83.14          \\
    Comet-QE               &                          & 74.92 / 57.48 / 80.63           & 77.01 / 58.64 / 81.69          & 77.66 / 59.15 / 82.25           & 77.85 / 59.42 / 82.59            & 78.18 / 59.60 / 82.83          \\
    Comet-Kiwi (sentence)  &                          & 75.20 / 57.31 / 80.26           & 78.05 / 58.50 / 81.23          & 79.56 / 59.24 / 81.81           & 80.36 / 59.65 / 82.20            & 80.81 / 59.91 / 82.49          \\
    Comet-QE (sentence)    &                          & 74.69 / 57.26 / 80.31           & 76.97 / 58.38 / 81.28          & 77.89 / 58.90 / 81.78           & 78.23 / 59.16 / 82.10            & 78.42 / 59.39 / 82.36          \\
    Doc-Comet-QE           &                          & 73.43 / 56.84 / 79.99           & 74.94 / 57.64 / 80.74          & 75.37 / 57.97 / 81.11           & 75.33 / 58.03 / 81.26            & 74.50 / 57.80 / 81.25          \\
    SLIDE (w=7, s=7)       &                          & 76.21 / 57.74 / 80.64           & 79.34 / 59.10 / 81.73          & 80.91 / 59.90 / 82.39           & 81.77 / \bb{60.40} / 82.85       & 82.27 / \bb{60.76} / \bc{83.18}\\
    SLIDE (w=7, s=1)       &                          & 76.23 / 57.71 / 80.64           & 79.41 / 59.12 / \bc{81.77}     & 80.89 / 59.89 / 82.41           & 81.77 / \bb{60.40} / \bc{82.87}  & 82.28 / 60.73 / \bc{83.18}     \\
    GEMBA-DA               &                          & \bg{76.37} / 57.46 / 80.37      & \bg{79.60} / 58.67 / 81.33     & \bg{81.12} / 59.34 / 81.83      & \bg{82.07} / 59.76 / 82.15       & \bg{82.64} / 60.06 / 82.36     \\
    EAPrompt               &                          & 74.30 / 56.99 / 79.86           & 76.62 / 57.97 / 80.57          & 77.79 / 58.53 / 80.87           & 78.35 / 58.84 / 81.00            & 78.41 / 58.97 / 80.93          \\
    EAPrompt (critical)    &                          & 74.66 / 57.04 / 79.98           & 77.07 / 57.98 / 80.68          & 78.24 / 58.41 / 80.99           & 78.64 / 58.52 / 81.08            & 78.68 / 58.42 / 81.00          \\
    \hline
    \end{tabular}
    \caption{QE reranking performance across all pool sizes reported as GEMBA-DA / BLEURT-20 / COMET-22 scores for each QE metric and translator. Best scores for each evaluator and pool size are highlighted and denoted in bold.}
    \label{tab:results}
\end{table*}
}
\label{tab:table 1}

\subsection{Pool Size}

We first examine the effect of candidate pool size on reranking performance. A pool size of one serves as the baseline, equivalent to no QE reranking. As shown in \autoref{fig:pool}, scores generally increase with larger pools, confirming the effectiveness of QE reranking at the document level. Gains are observed for both LLMs and NMT models, with the largest improvements in Yaraku Translate, likely reflecting its document-level training. Performance does not reach a full plateau at pool size 32, suggesting larger pools could yield further gains. While improvements are consistent across all evaluators, the leading QE metric varies, likely due to evaluator-specific biases such as overfitting (e.g., COMET-based QE metrics evaluated with COMET-22), and sequence length constraints in COMET-22 and BLEURT, which limit their ability to fully capture the advantages of LLM-based QE metrics which can handle longer sequences. Full results are given in \autoref{tab:table 1}.

Among COMET-based metrics, Comet-Kiwi consistently outperforms COMET-QE. Notably, scoring entire documents in one pass also outperforms per-sentence averaging in all settings. The gap is smallest, however, for Yaraku Translate, which enforces sentence alignment, suggesting the benefit of aligned outputs. In contrast, Doc-COMET-QE provides no benefit and ranks among the weakest QE metrics. Both configurations of SLIDE performed very similarly to each other and to standard Comet-Kiwi, only providing marginal gains in some cases. This is expected, as our SLIDE implementation is built upon Comet-Kiwi. 

The LLM-based metrics show mixed results. When evaluated with BLEURT and COMET, GEMBA-DA performs well but slightly below the best-performing COMET-based metrics. This is unsurprising due to the potential overfitting risks of the COMET-based QE metrics. However, with GPT as the evaluator, GEMBA-DA achieves the best performance. Naturally, there is also an overfitting risk in this case, but the difference in backbone LLM attempts to minimise this. Surprisingly, while adding the critical category helps, EAPrompt generally showed poor performance.

\subsection{Length}

\begin{figure}[t]
    \centering
    \includegraphics[width=\linewidth]{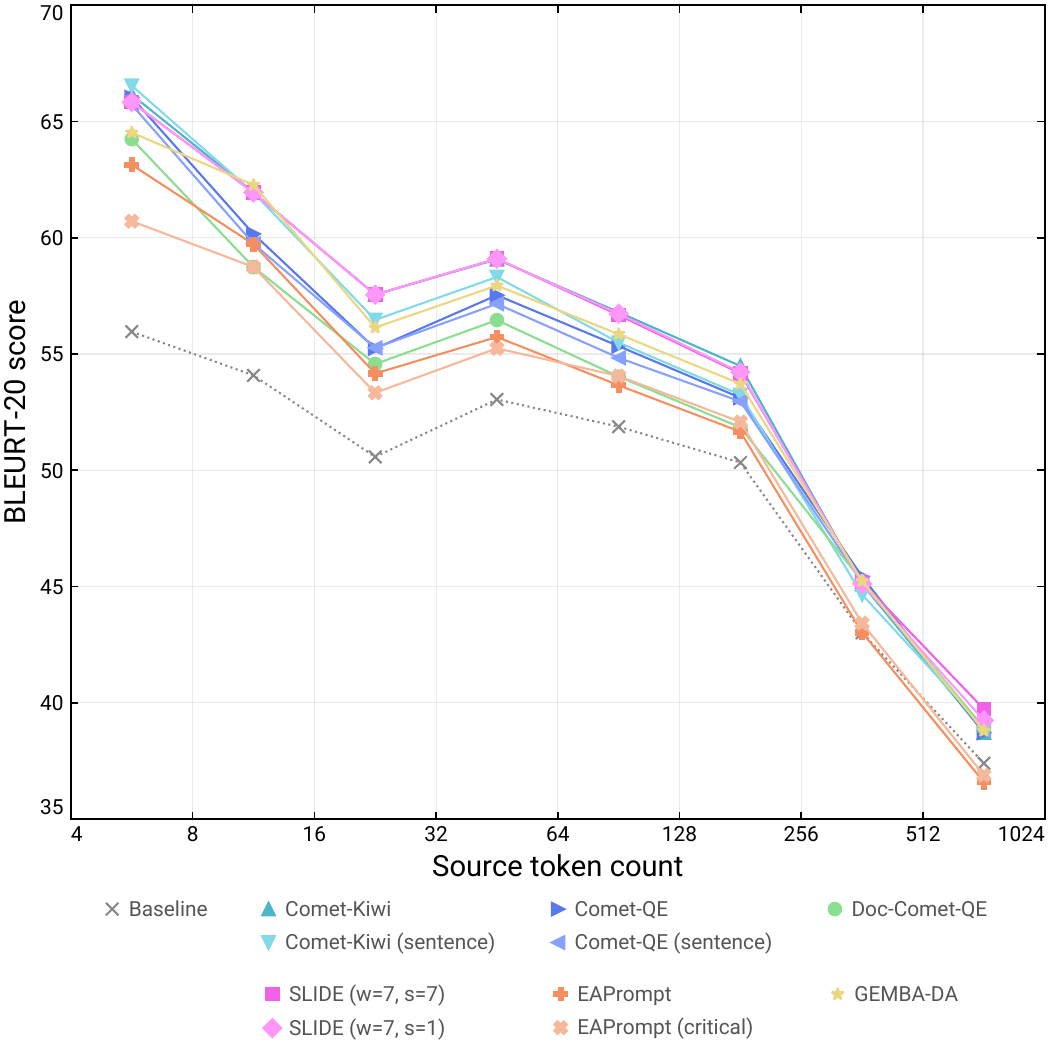}
    \caption{QE reranking performance for all QE metrics at pool size 32, averaged across all translator models. Gains diminish with longer documents but remain above the baseline (pool size 1) for most metrics.}
    \label{fig:length}
\end{figure}

We expected QE performance to degrade with longer inputs: learned metrics were not trained for long sequences, and LLM attention tends to diverge over extended contexts \cite{barbero2025why}. The results shown in \autoref{fig:length} confirm this hypothesis. QE metrics perform best on short inputs, but performance remains stable up to ~256 source tokens, indicating reasonable capability in multi-sentence contexts. Beyond this point, performance rapidly declines, reflecting the 512 token limit for combined source and target texts in most QE metrics and evaluators. Nonetheless, most QE metrics continue to provide a performance gain over the baseline even at the longest tested sequence lengths.

The extent of degradation varies across metrics. SLIDE exhibits similar performance to Comet-Kiwi for short sequences but retains slightly higher performance for long sequences, thanks to the sliding window approach avoiding the token limit. Additionally, when GPT is used as an evaluator, the lead of GEMBA-DA as a QE metric becomes the biggest for long sequences, highlighting the long-context capabilities of LLMs.

\subsection{Runtime}

\begin{figure}[t]
    \centering
    \includegraphics[width=\linewidth]{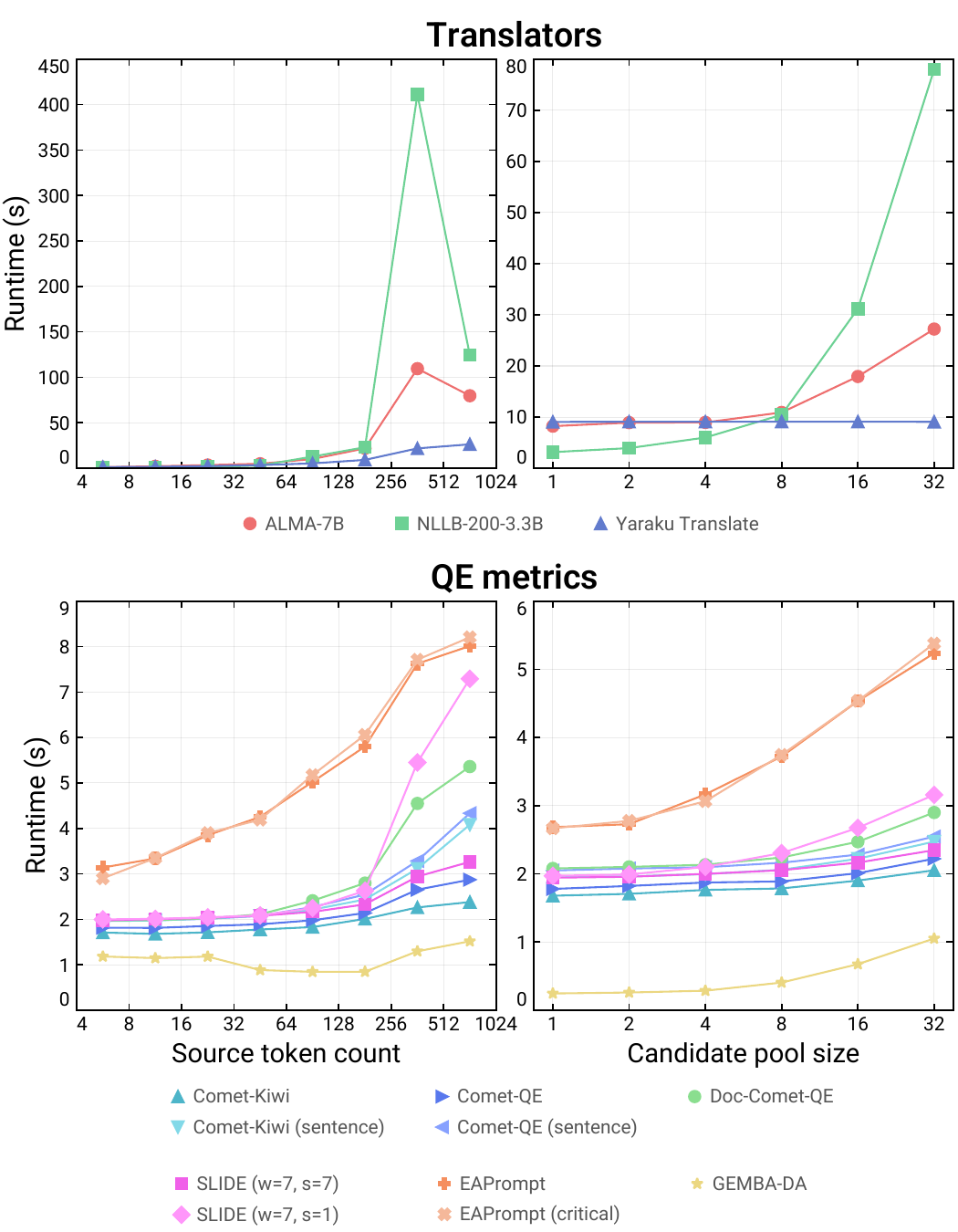}
    \caption{Runtime by source length and pool size for all QE metrics and translators. Translation runtime rises steeply for models not trained at the document level, while QE runtime remains a small fraction of the overall runtime.}
    \label{fig:runtime}
\end{figure}

Runtime requirements are difficult to assess in a hardware-agnostic manner, as they depend heavily on GPU memory, implementation details, batch size, and document length. In our experiments, we used a cluster of 4 NVIDIA A6000 GPUs, sharding models across devices so that the maximum pool size could be processed in a single batch. To keep comparison relatively fair, we fixed the batch size for both translation and QE models at 32 (the largest pool size). While many QE models could support larger batch sizes, this cap highlights the potential slowdown of metrics requiring multiple evaluations per candidate. We show the runtime across translation models and QE metrics in \autoref{fig:runtime}.

Among translators, Yaraku Translate shows little sensitivity to pool size, likely reflecting the efficiency of its dynamic beam search decoding. Counterintuitively, the smaller NLLB model is slower than the larger ALMA, exploding exponentially with both pool size and input length. This is primarily due to difficulty generating stop tokens; hallucinated outputs often reach the token limit, delaying the entire batch. This highlights the value of document-level translation models and effective stopping strategies. To partially mitigate this issue, we apply an adaptive maximum token limit defined as:

\[
\min\!\left( 
  N_{\text{ceil}},\;
  \left\lceil 
    L_{\text{in}} \cdot \alpha_{m} \cdot \frac{\mu_{\text{tgt}}}{\mu_{\text{src}}} + \alpha_{a} 
  \right\rceil
\right)
\]

where $L_{\mathrm{in}}$ is the input token length, $\alpha_{a}=10$, $\alpha_{m}=2$ are additive and multiplicative margin factors, $N_{\mathrm{ceil}}=2048$ is a hard ceiling, and $\mu_{\mathrm{tgt}}$, $\mu_{\mathrm{src}}$ denote the average dataset token lengths for the current target and source languages, respectively. This caps hallucinated translations while retaining sufficient headroom for legitimate document-length variability.

For learned QE metrics, runtime grows modestly with pool size and sequence length. Methods requiring multiple evaluations per candidate, namely SLIDE, Doc-COMET-QE, and the sentence-based COMET variants, exhibit steeper growth on longer documents. Between SLIDE configurations, while $s=1$ and $s=7$ exhibit similar performance, $s=1$ incurs significantly higher runtime, making $s=7$ the more practical choice. Comparison between learned QE metrics and LLM-based methods is complicated by experiment setup limitations (Gemma was hosted on a GH200 NVIDIA GPU), yet GEMBA-DA runs substantially faster than both EAPrompt variants, as it requires minimal token generation. GEMBA-DA, however, exhibits higher runtime growth with larger pool sizes, likely reflecting the higher probability of triggering its failure-prevention strategy.

Overall, QE runtime represents a small fraction of translator runtime, making reranking potentially near cost-free when using efficient translator models, provided hardware is otherwise not being used to simultaneously translate multiple distinct inputs, and the largest pool size fits into a single batch. Beyond this threshold, larger pools represent a trade-off between additional translation quality and runtime cost.

\section{Conclusion}
We investigated the applicability of QE reranking to the document translation domain and found consistent translation quality gains over standard decoding across various QE metrics and translation models. Gains increased with translation candidate pool size and were not saturated at 32, indicating further potential improvement. Methods that score full documents in one pass consistently outperform sentence-level averaging, even with QE metrics designed for sentence-level scoring. SLIDE was found to be the best performing QE metric, matching Comet-Kiwi on short inputs while being more performant on long documents. Among LLM-based methods, GEMBA-DA was found to be competitive when evaluated with COMET-22 and BLEURT, and leads under GPT evaluation. Although performance gains decline for long documents beyond 256 source tokens, QE reranking improves translation quality even for the longest documents in our dataset. Runtime analysis showed that all QE metrics represent a fraction of total translation runtime cost, allowing for near cost-free performance gains under certain conditions.

\section*{Limitations}
This study has several limitations. First, although we tested multiple pool sizes, performance did not reach a clear saturation point (i.e., a peak followed by stagnation), which would have provided stronger evidence of the limits of QE reranking. Second, resource constraints prevented us from exploring more diverse LLM prompting methods for evaluation, limiting our ability to fully exploit the potential of LLMs. Third, most QE models remain constrained by a 512-token limit, restricting their applicability for longer documents. Lastly, this study lacks human evaluation, which reduces the reliability of our tested evaluation metrics, and would have allowed us to better explore the extent of overfitting.

\bibliography{custom}

\begin{thebibliography}{34}
\expandafter\ifx\csname natexlab\endcsname\relax\def\natexlab#1{#1}\fi

\bibitem[{Barbero et~al.(2025)Barbero, Arroyo, Gu, Perivolaropoulos, Veli{\v{c}}kovi{\'c}, Pascanu, and Bronstein}]{barbero2025why}
Federico Barbero, Alvaro Arroyo, Xiangming Gu, Christos Perivolaropoulos, Petar Veli{\v{c}}kovi{\'c}, Razvan Pascanu, and Michael~M. Bronstein. 2025.
\newblock \href {https://openreview.net/forum?id=tu4dFUsW5z} {Why do {LLM}s attend to the first token?}
\newblock In \emph{Second Conference on Language Modeling}.

\bibitem[{Chi et~al.(2021)Chi, Dong, Wei, Yang, Singhal, Wang, Song, Mao, Huang, and Zhou}]{chi-etal-2021-infoxlm}
Zewen Chi, Li~Dong, Furu Wei, Nan Yang, Saksham Singhal, Wenhui Wang, Xia Song, Xian-Ling Mao, Heyan Huang, and Ming Zhou. 2021.
\newblock \href {https://doi.org/10.18653/v1/2021.naacl-main.280} {{I}nfo{XLM}: An information-theoretic framework for cross-lingual language model pre-training}.
\newblock In \emph{Proceedings of the 2021 Conference of the North American Chapter of the Association for Computational Linguistics: Human Language Technologies}, pages 3576--3588, Online. Association for Computational Linguistics.

\bibitem[{Deutsch et~al.(2023)Deutsch, Juraska, Finkelstein, and Freitag}]{deutsch-etal-2023-training}
Daniel Deutsch, Juraj Juraska, Mara Finkelstein, and Markus Freitag. 2023.
\newblock \href {https://doi.org/10.18653/v1/2023.wmt-1.96} {Training and meta-evaluating machine translation evaluation metrics at the paragraph level}.
\newblock In \emph{Proceedings of the Eighth Conference on Machine Translation}, pages 996--1013, Singapore. Association for Computational Linguistics.

\bibitem[{Eikema and Aziz(2020)}]{eikema-aziz-2020-map}
Bryan Eikema and Wilker Aziz. 2020.
\newblock \href {https://doi.org/10.18653/v1/2020.coling-main.398} {Is {MAP} decoding all you need? the inadequacy of the mode in neural machine translation}.
\newblock In \emph{Proceedings of the 28th International Conference on Computational Linguistics}, pages 4506--4520, Barcelona, Spain (Online). International Committee on Computational Linguistics.

\bibitem[{Fernandes et~al.(2022)Fernandes, Farinhas, Rei, C.~de Souza, Ogayo, Neubig, and Martins}]{fernandes-etal-2022-quality}
Patrick Fernandes, Ant{\'o}nio Farinhas, Ricardo Rei, Jos{\'e}~G. C.~de Souza, Perez Ogayo, Graham Neubig, and Andre Martins. 2022.
\newblock \href {https://doi.org/10.18653/v1/2022.naacl-main.100} {Quality-aware decoding for neural machine translation}.
\newblock In \emph{Proceedings of the 2022 Conference of the North American Chapter of the Association for Computational Linguistics: Human Language Technologies}, pages 1396--1412, Seattle, United States. Association for Computational Linguistics.

\bibitem[{Freitag et~al.(2021)Freitag, Foster, Grangier, Ratnakar, Tan, and Macherey}]{freitag-etal-2021-experts}
Markus Freitag, George Foster, David Grangier, Viresh Ratnakar, Qijun Tan, and Wolfgang Macherey. 2021.
\newblock \href {https://doi.org/10.1162/tacl_a_00437} {Experts, errors, and context: A large-scale study of human evaluation for machine translation}.
\newblock \emph{Transactions of the Association for Computational Linguistics}, 9:1460--1474.

\bibitem[{Freitag et~al.(2023{\natexlab{a}})Freitag, Ghorbani, and Fernandes}]{freitag2023epsilonsamplingrocksinvestigating}
Markus Freitag, Behrooz Ghorbani, and Patrick Fernandes. 2023{\natexlab{a}}.
\newblock \href {http://arxiv.org/abs/2305.09860} {Epsilon sampling rocks: Investigating sampling strategies for minimum bayes risk decoding for machine translation}.

\bibitem[{Freitag et~al.(2022{\natexlab{a}})Freitag, Grangier, Tan, and Liang}]{freitag2022highqualityhighmodel}
Markus Freitag, David Grangier, Qijun Tan, and Bowen Liang. 2022{\natexlab{a}}.
\newblock \href {http://arxiv.org/abs/2111.09388} {High quality rather than high model probability: Minimum bayes risk decoding with neural metrics}.

\bibitem[{Freitag et~al.(2023{\natexlab{b}})Freitag, Mathur, Lo, Avramidis, Rei, Thompson, Kocmi, Blain, Deutsch, Stewart, Zerva, Castilho, Lavie, and Foster}]{freitag-etal-2023-results}
Markus Freitag, Nitika Mathur, Chi-kiu Lo, Eleftherios Avramidis, Ricardo Rei, Brian Thompson, Tom Kocmi, Frederic Blain, Daniel Deutsch, Craig Stewart, Chrysoula Zerva, Sheila Castilho, Alon Lavie, and George Foster. 2023{\natexlab{b}}.
\newblock \href {https://doi.org/10.18653/v1/2023.wmt-1.51} {Results of {WMT}23 metrics shared task: Metrics might be guilty but references are not innocent}.
\newblock In \emph{Proceedings of the Eighth Conference on Machine Translation}, pages 578--628, Singapore. Association for Computational Linguistics.

\bibitem[{Freitag et~al.(2022{\natexlab{b}})Freitag, Rei, Mathur, Lo, Stewart, Avramidis, Kocmi, Foster, Lavie, and Martins}]{freitag-etal-2022-results}
Markus Freitag, Ricardo Rei, Nitika Mathur, Chi-kiu Lo, Craig Stewart, Eleftherios Avramidis, Tom Kocmi, George Foster, Alon Lavie, and Andr{\'e} F.~T. Martins. 2022{\natexlab{b}}.
\newblock \href {https://aclanthology.org/2022.wmt-1.2/} {Results of {WMT}22 metrics shared task: Stop using {BLEU} {--} neural metrics are better and more robust}.
\newblock In \emph{Proceedings of the Seventh Conference on Machine Translation (WMT)}, pages 46--68, Abu Dhabi, United Arab Emirates (Hybrid). Association for Computational Linguistics.

\bibitem[{Karpinska and Iyyer(2023)}]{karpinska-iyyer-2023-large}
Marzena Karpinska and Mohit Iyyer. 2023.
\newblock \href {https://doi.org/10.18653/v1/2023.wmt-1.41} {Large language models effectively leverage document-level context for literary translation, but critical errors persist}.
\newblock In \emph{Proceedings of the Eighth Conference on Machine Translation}, pages 419--451, Singapore. Association for Computational Linguistics.

\bibitem[{Kiss and Strunk(2006)}]{kiss-strunk-2006-unsupervised}
Tibor Kiss and Jan Strunk. 2006.
\newblock \href {https://doi.org/10.1162/coli.2006.32.4.485} {Unsupervised multilingual sentence boundary detection}.
\newblock \emph{Computational Linguistics}, 32(4):485--525.

\bibitem[{Kocmi and Federmann(2023)}]{kocmi-federmann-2023-large}
Tom Kocmi and Christian Federmann. 2023.
\newblock \href {https://aclanthology.org/2023.eamt-1.19/} {Large language models are state-of-the-art evaluators of translation quality}.
\newblock In \emph{Proceedings of the 24th Annual Conference of the European Association for Machine Translation}, pages 193--203, Tampere, Finland. European Association for Machine Translation.

\bibitem[{Kudo et~al.(2024)Kudo, Deguchi, Morishita, Fujii, Ito, Ozaki, Natsumi, Sato, Yano, Takahashi, Kimura, Hara, Sakai, and Suzuki}]{kudo-etal-2024-document}
Keito Kudo, Hiroyuki Deguchi, Makoto Morishita, Ryo Fujii, Takumi Ito, Shintaro Ozaki, Koki Natsumi, Kai Sato, Kazuki Yano, Ryosuke Takahashi, Subaru Kimura, Tomomasa Hara, Yusuke Sakai, and Jun Suzuki. 2024.
\newblock \href {https://doi.org/10.18653/v1/2024.wmt-1.14} {Document-level translation with {LLM} reranking: Team-{J} at {WMT} 2024 general translation task}.
\newblock In \emph{Proceedings of the Ninth Conference on Machine Translation}, pages 210--226, Miami, Florida, USA. Association for Computational Linguistics.

\bibitem[{Kumar and Byrne(2004)}]{kumar-byrne-2004-minimum}
Shankar Kumar and William Byrne. 2004.
\newblock \href {https://aclanthology.org/N04-1022/} {Minimum {B}ayes-risk decoding for statistical machine translation}.
\newblock In \emph{Proceedings of the Human Language Technology Conference of the North {A}merican Chapter of the Association for Computational Linguistics: {HLT}-{NAACL} 2004}, pages 169--176, Boston, Massachusetts, USA. Association for Computational Linguistics.

\bibitem[{Lu et~al.(2024)Lu, Qiu, Ding, Zhang, Kocmi, and Tao}]{lu-etal-2024-error}
Qingyu Lu, Baopu Qiu, Liang Ding, Kanjian Zhang, Tom Kocmi, and Dacheng Tao. 2024.
\newblock \href {https://doi.org/10.18653/v1/2024.findings-acl.520} {Error analysis prompting enables human-like translation evaluation in large language models}.
\newblock In \emph{Findings of the Association for Computational Linguistics: ACL 2024}, pages 8801--8816, Bangkok, Thailand. Association for Computational Linguistics.

\bibitem[{Ma et~al.(2019)Ma, Wei, Bojar, and Graham}]{ma-etal-2019-results}
Qingsong Ma, Johnny Wei, Ond{\v{r}}ej Bojar, and Yvette Graham. 2019.
\newblock \href {https://doi.org/10.18653/v1/W19-5302} {Results of the {WMT}19 metrics shared task: Segment-level and strong {MT} systems pose big challenges}.
\newblock In \emph{Proceedings of the Fourth Conference on Machine Translation (Volume 2: Shared Task Papers, Day 1)}, pages 62--90, Florence, Italy. Association for Computational Linguistics.

\bibitem[{M{\"u}ller and Sennrich(2021)}]{muller-sennrich-2021-understanding}
Mathias M{\"u}ller and Rico Sennrich. 2021.
\newblock \href {https://doi.org/10.18653/v1/2021.acl-long.22} {Understanding the properties of minimum {B}ayes risk decoding in neural machine translation}.
\newblock In \emph{Proceedings of the 59th Annual Meeting of the Association for Computational Linguistics and the 11th International Joint Conference on Natural Language Processing (Volume 1: Long Papers)}, pages 259--272, Online. Association for Computational Linguistics.

\bibitem[{OpenAI(2024)}]{openai2024gpt4technicalreport}
OpenAI. 2024.
\newblock \href {http://arxiv.org/abs/2303.08774} {Gpt-4 technical report}.

\bibitem[{Papineni et~al.(2002)Papineni, Roukos, Ward, and Zhu}]{papineni-etal-2002-bleu}
Kishore Papineni, Salim Roukos, Todd Ward, and Wei-Jing Zhu. 2002.
\newblock \href {https://doi.org/10.3115/1073083.1073135} {{B}leu: a method for automatic evaluation of machine translation}.
\newblock In \emph{Proceedings of the 40th Annual Meeting of the Association for Computational Linguistics}, pages 311--318, Philadelphia, Pennsylvania, USA. Association for Computational Linguistics.

\bibitem[{Raunak et~al.(2024)Raunak, Kocmi, and Post}]{raunak-etal-2024-slide}
Vikas Raunak, Tom Kocmi, and Matt Post. 2024.
\newblock \href {https://doi.org/10.18653/v1/2024.naacl-short.18} {{SLIDE}: Reference-free evaluation for machine translation using a sliding document window}.
\newblock In \emph{Proceedings of the 2024 Conference of the North American Chapter of the Association for Computational Linguistics: Human Language Technologies (Volume 2: Short Papers)}, pages 205--211, Mexico City, Mexico. Association for Computational Linguistics.

\bibitem[{Rei et~al.(2022{\natexlab{a}})Rei, C.~de Souza, Alves, Zerva, Farinha, Glushkova, Lavie, Coheur, and Martins}]{rei-etal-2022-comet}
Ricardo Rei, Jos{\'e}~G. C.~de Souza, Duarte Alves, Chrysoula Zerva, Ana~C Farinha, Taisiya Glushkova, Alon Lavie, Luisa Coheur, and Andr{\'e} F.~T. Martins. 2022{\natexlab{a}}.
\newblock \href {https://aclanthology.org/2022.wmt-1.52/} {{COMET}-22: Unbabel-{IST} 2022 submission for the metrics shared task}.
\newblock In \emph{Proceedings of the Seventh Conference on Machine Translation (WMT)}, pages 578--585, Abu Dhabi, United Arab Emirates (Hybrid). Association for Computational Linguistics.

\bibitem[{Rei et~al.(2020{\natexlab{a}})Rei, Stewart, Farinha, and Lavie}]{rei-etal-2020-comet}
Ricardo Rei, Craig Stewart, Ana~C Farinha, and Alon Lavie. 2020{\natexlab{a}}.
\newblock \href {https://doi.org/10.18653/v1/2020.emnlp-main.213} {{COMET}: A neural framework for {MT} evaluation}.
\newblock In \emph{Proceedings of the 2020 Conference on Empirical Methods in Natural Language Processing (EMNLP)}, pages 2685--2702, Online. Association for Computational Linguistics.

\bibitem[{Rei et~al.(2020{\natexlab{b}})Rei, Stewart, Farinha, and Lavie}]{rei-etal-2020-unbabels}
Ricardo Rei, Craig Stewart, Ana~C Farinha, and Alon Lavie. 2020{\natexlab{b}}.
\newblock \href {https://aclanthology.org/2020.wmt-1.101/} {Unbabel{'}s participation in the {WMT}20 metrics shared task}.
\newblock In \emph{Proceedings of the Fifth Conference on Machine Translation}, pages 911--920, Online. Association for Computational Linguistics.

\bibitem[{Rei et~al.(2022{\natexlab{b}})Rei, Treviso, Guerreiro, Zerva, Farinha, Maroti, C.~de Souza, Glushkova, Alves, Coheur, Lavie, and Martins}]{rei-etal-2022-cometkiwi}
Ricardo Rei, Marcos Treviso, Nuno~M. Guerreiro, Chrysoula Zerva, Ana~C Farinha, Christine Maroti, Jos{\'e}~G. C.~de Souza, Taisiya Glushkova, Duarte Alves, Luisa Coheur, Alon Lavie, and Andr{\'e} F.~T. Martins. 2022{\natexlab{b}}.
\newblock \href {https://aclanthology.org/2022.wmt-1.60/} {{C}omet{K}iwi: {IST}-unbabel 2022 submission for the quality estimation shared task}.
\newblock In \emph{Proceedings of the Seventh Conference on Machine Translation (WMT)}, pages 634--645, Abu Dhabi, United Arab Emirates (Hybrid). Association for Computational Linguistics.

\bibitem[{Sellam et~al.(2020)Sellam, Das, and Parikh}]{sellam-etal-2020-bleurt}
Thibault Sellam, Dipanjan Das, and Ankur Parikh. 2020.
\newblock \href {https://doi.org/10.18653/v1/2020.acl-main.704} {{BLEURT}: Learning robust metrics for text generation}.
\newblock In \emph{Proceedings of the 58th Annual Meeting of the Association for Computational Linguistics}, pages 7881--7892, Online. Association for Computational Linguistics.

\bibitem[{Team(2025)}]{gemmateam2025gemma3technicalreport}
Gemma Team. 2025.
\newblock \href {http://arxiv.org/abs/2503.19786} {Gemma 3 technical report}.

\bibitem[{Team(2022)}]{nllbteam2022languageleftbehindscaling}
NLLB Team. 2022.
\newblock \href {http://arxiv.org/abs/2207.04672} {No language left behind: Scaling human-centered machine translation}.

\bibitem[{Touvron et~al.(2023)Touvron, Martin, Stone, Albert, Almahairi, Babaei, Bashlykov, Batra, Bhargava, Bhosale, Bikel, Blecher, Ferrer, Chen, Cucurull, Esiobu, Fernandes, Fu, Fu, Fuller, Gao, Goswami, Goyal, Hartshorn, Hosseini, Hou, Inan, Kardas, Kerkez, Khabsa, Kloumann, Korenev, Koura, Lachaux, Lavril, Lee, Liskovich, Lu, Mao, Martinet, Mihaylov, Mishra, Molybog, Nie, Poulton, Reizenstein, Rungta, Saladi, Schelten, Silva, Smith, Subramanian, Tan, Tang, Taylor, Williams, Kuan, Xu, Yan, Zarov, Zhang, Fan, Kambadur, Narang, Rodriguez, Stojnic, Edunov, and Scialom}]{touvron2023llama2openfoundation}
Hugo Touvron, Louis Martin, Kevin Stone, Peter Albert, Amjad Almahairi, Yasmine Babaei, Nikolay Bashlykov, Soumya Batra, Prajjwal Bhargava, Shruti Bhosale, Dan Bikel, Lukas Blecher, Cristian~Canton Ferrer, Moya Chen, Guillem Cucurull, David Esiobu, Jude Fernandes, Jeremy Fu, Wenyin Fu, Brian Fuller, Cynthia Gao, Vedanuj Goswami, Naman Goyal, Anthony Hartshorn, Saghar Hosseini, Rui Hou, Hakan Inan, Marcin Kardas, Viktor Kerkez, Madian Khabsa, Isabel Kloumann, Artem Korenev, Punit~Singh Koura, Marie-Anne Lachaux, Thibaut Lavril, Jenya Lee, Diana Liskovich, Yinghai Lu, Yuning Mao, Xavier Martinet, Todor Mihaylov, Pushkar Mishra, Igor Molybog, Yixin Nie, Andrew Poulton, Jeremy Reizenstein, Rashi Rungta, Kalyan Saladi, Alan Schelten, Ruan Silva, Eric~Michael Smith, Ranjan Subramanian, Xiaoqing~Ellen Tan, Binh Tang, Ross Taylor, Adina Williams, Jian~Xiang Kuan, Puxin Xu, Zheng Yan, Iliyan Zarov, Yuchen Zhang, Angela Fan, Melanie Kambadur, Sharan Narang, Aurelien Rodriguez, Robert Stojnic, Sergey Edunov, and Thomas
  Scialom. 2023.
\newblock \href {http://arxiv.org/abs/2307.09288} {Llama 2: Open foundation and fine-tuned chat models}.

\bibitem[{Vernikos and Popescu-Belis(2024)}]{vernikos-popescu-belis-2024-dont}
Giorgos Vernikos and Andrei Popescu-Belis. 2024.
\newblock \href {https://doi.org/10.18653/v1/2024.acl-long.653} {Don{'}t rank, combine! combining machine translation hypotheses using quality estimation}.
\newblock In \emph{Proceedings of the 62nd Annual Meeting of the Association for Computational Linguistics (Volume 1: Long Papers)}, pages 12087--12105, Bangkok, Thailand. Association for Computational Linguistics.

\bibitem[{Vernikos et~al.(2022)Vernikos, Thompson, Mathur, and Federico}]{vernikos-etal-2022-embarrassingly}
Giorgos Vernikos, Brian Thompson, Prashant Mathur, and Marcello Federico. 2022.
\newblock \href {https://aclanthology.org/2022.wmt-1.6/} {Embarrassingly easy document-level {MT} metrics: How to convert any pretrained metric into a document-level metric}.
\newblock In \emph{Proceedings of the Seventh Conference on Machine Translation (WMT)}, pages 118--128, Abu Dhabi, United Arab Emirates (Hybrid). Association for Computational Linguistics.

\bibitem[{Vijayakumar et~al.(2018)Vijayakumar, Cogswell, Selvaraju, Sun, Lee, Crandall, and Batra}]{vijayakumar2018diversebeamsearchdecoding}
Ashwin~K Vijayakumar, Michael Cogswell, Ramprasath~R. Selvaraju, Qing Sun, Stefan Lee, David Crandall, and Dhruv Batra. 2018.
\newblock \href {http://arxiv.org/abs/1610.02424} {Diverse beam search: Decoding diverse solutions from neural sequence models}.

\bibitem[{Wang et~al.(2024)Wang, Du, Jiao, Lyu, Pang, Cui, Song, Wong, Shi, and Tu}]{wang-etal-2024-benchmarking}
Longyue Wang, Zefeng Du, Wenxiang Jiao, Chenyang Lyu, Jianhui Pang, Leyang Cui, Kaiqiang Song, Derek Wong, Shuming Shi, and Zhaopeng Tu. 2024.
\newblock \href {https://doi.org/10.18653/v1/2024.findings-acl.428} {Benchmarking and improving long-text translation with large language models}.
\newblock In \emph{Findings of the Association for Computational Linguistics: ACL 2024}, pages 7175--7187, Bangkok, Thailand. Association for Computational Linguistics.

\bibitem[{Xu et~al.(2024)Xu, Kim, Sharaf, and Awadalla}]{xu2024paradigmshiftmachinetranslation}
Haoran Xu, Young~Jin Kim, Amr Sharaf, and Hany~Hassan Awadalla. 2024.
\newblock \href {http://arxiv.org/abs/2309.11674} {A paradigm shift in machine translation: Boosting translation performance of large language models}.

\end{thebibliography}

\end{document}